\documentclass[sigconf]{acmart}
\AtBeginDocument{%
  }

\setcopyright{acmlicensed}
\copyrightyear{2026}
\acmYear{2026}
\acmDOI{XXXXXXX.XXXXXXX}
\usepackage{hyperref}
\usepackage{url}

\usepackage{algorithm}
\usepackage[noend]{algpseudocode}
\usepackage[utf8]{inputenc} % allow utf-8 input
\usepackage[T1]{fontenc}    % use 8-bit T1 fonts
\usepackage{hyperref}       % hyperlinks
\usepackage{url}            % simple URL typesetting
\usepackage{booktabs}       % professional-quality tables
\usepackage{amsfonts}       % blackboard math symbols
\usepackage{microtype}      % microtypography
\usepackage{xcolor}         % colors
\usepackage{balance}
\usepackage{amsmath} 
\usepackage{multirow}
\usepackage{tabularx}
\usepackage{booktabs}
\usepackage{graphicx} 
\usepackage{colortbl}
\usepackage{makecell}
\usepackage{arydshln}
\usepackage{wrapfig}
\usepackage{caption}
\usepackage{subcaption}
\usepackage{xcolor}
\usepackage{mathrsfs}
\usepackage[most]{tcolorbox}
\definecolor{officeblue}{RGB}{79,129,189}   % 深雅蓝
\definecolor{officegreen}{RGB}{155,187,89}  % 柔和绿
\definecolor{officeorange}{RGB}{217,150,148} % 雅致橙红

\tcbset{
  promptbox/.style={
    enhanced,
    colback=white,
    colframe=black!15,
    boxrule=0.6pt,
    arc=3pt,
    left=8pt, right=8pt, top=8pt, bottom=8pt,
    title style={font=\bfseries},
    fontupper=\normalsize\rmfamily,
  },
}

\begin{document}
\def\method{ConceptFormer}
\title{\method{}: Learning Adaptive Latent Concepts for Query-Document Alignment in Visual Document Retrieval}% 
\author{Chunyi Peng}
% \authornote{ \ \ indicates equal contribution.}
\affiliation{%
  \institution{Northeastern University}
  \city{Shenyang}
  \country{China}
}
\email{hm.cypeng@gmail.com}

\author{Zhipeng Xu}
\affiliation{%
  \institution{Northeastern University}
  \city{Shenyang}
  \country{China}
}
\email{xuzp@mails.neu.edu.cn}

\author{Yukun Yan}
\authornote{ \ \ indicates corresponding author.}
\affiliation{%
  \institution{Tsinghua University}
  \city{Beijing}
  \country{China}
}
\email{yanyk.thu@gmail.com}

\author{Zhenghao Liu}
\authornotemark[1]
\affiliation{%
  \institution{Northeastern University}
  \city{Shenyang}
  \country{China}}
\email{liuzhenghao@mail.neu.edu.cn}

\author{Shi Yu}
\affiliation{
  \institution{Tsinghua University}
  \city{Beijing}
  \country{China}
}
\email{yushi17@foxmail.com}

\author{Sen Mei}
\affiliation{
  \institution{Tsinghua University}
  \city{Beijing}
  \country{China}
}
\email{meisen2025@gmail.com}

\author{Yubo Sun}
\affiliation{
  \institution{Peking University}
  \city{Beijing}
  \country{China}
}
\email{boggysyb@gmail.com}

\author{Yongheng Zhang}
\affiliation{%
  \institution{Tsinghua University}
  \city{Beijing}
  \country{China}
}
\email{zyhbrz@gmail.com}

\author{Jie Zhou}
\affiliation{%
  \institution{Tsinghua University}
  \city{Beijing}
  \country{China}
}
\email{zhoujie19940804@126.com}

\author{Yu Gu}
\affiliation{%
  \institution{Northeastern University}
  \city{Shenyang}
  \country{China}}
\email{guyu@mail.neu.edu.cn}

\author{Ge Yu}
\affiliation{%
  \institution{Northeastern University}
  \city{Shenyang}
  \country{China}}
\email{yuge@mail.neu.edu.cn}

\author{Maosong Sun}
\affiliation{%
  \institution{Tsinghua University}
  \city{Beijing}
  \country{China}
}
\email{sms@tsinghua.edu.cn}

\renewcommand{\shortauthors}{Chunyi Peng et al.}

% % \ccsdesc[300]{Do Not Use This Code~Generate the Correct Terms for Your Paper}
% % % \ccsdesc{Do Not Use This Code~Generate the Correct Terms for Your Paper}
% % \ccsdesc[100]{Do Not Use This Code~Generate the Correct Terms for Your Paper}

%\iclrfinalcopy % Uncomment for camera-ready version, but NOT for submission.
\begin{abstract}
Visual document retrieval is a critical component of multimodal retrieval-augmented generation, aiming to identify query-relevant pages from document collections where evidence is distributed across text, layout, charts, and visual structures. Existing visual document retrievers typically encode full-page images and optimize query-page relevance through page-level contrastive learning. While effective, such supervision provides limited guidance on which localized evidence contributes to the relevance between a query and a page. Recent efforts toward finer-grained supervision primarily rely on textual descriptions or localized visual regions as evidence proxies. However, such supervision signals may either overlook complex visual structures or provide incomplete and inaccurate representations of the underlying evidence.
To address these limitations, we propose \textbf{\method{}}, a latent concept representation learning framework for visual document retrieval. \method{} models query-relevant evidence as continuous, query-conditioned latent concepts that explicitly bridge localized visual evidence and semantic relevance, without requiring either textual intermediate representations or direct reliance on raw visual annotations. During training, \method{} employs a strong vision-language model to dynamically determine the number of latent concept tokens and uses these concepts as an intermediate representation to bridge the semantic gap between queries and documents, thereby guiding the learning of the embedding space.
Experiments on diverse visual document retrieval benchmarks demonstrate that \method{} achieves 16.7\% and 22.1\% relative improvements in average NDCG@10 over the strongest visual retrieval baseline and the strongest OCR-based text retrieval baseline, respectively. Further analysis reveals that latent concepts effectively connect localized visual evidence with semantic relevance, enabling the retriever to capture both fine-grained textual cues and complex document-level visual structures while preserving strong retrieval alignment. Codes and data are available at \url{https://github.com/Neuir/ConceptFormer}.

\end{abstract}

\maketitle

\section{Introduction}

Vision-language models (VLMs)~\cite{qwen36plus, bai2025qwen2, abdin2024phi} have demonstrated remarkable perception and generation capabilities across document question answering~\cite{mathew2021docvqa, antol2015vqa}, information extraction~\cite{tang2023unifying, xu2026cc}, and multimodal reasoning~\cite{peng2026mixture, chen2026towards}.
To preserve the layout-dependent structures and rich visual semantics inherent in visual documents, recent studies~\cite{sun2025visrag, wang2026vrag, xiong2026lang2act} address these tasks by moving beyond plain-text serialization-based approaches, which may introduce information loss and cascading errors~\cite{yu2025visrag}. Instead, they directly process document images using full-page screenshots together with zoom-in crops or region-level visual reasoning to retain both global page context and fine-grained visual evidence.
Among these applications, visual document retrieval~\cite{marinai2011digital, doermann1998indexing, alaei2016brief} serves as a fundamental component of multimodal document understanding pipelines, aiming to identify query-relevant documents from large-scale collections. However, existing retrieval systems remain constrained by the ability to accurately capture fine-grained query-document relevance, limiting their effectiveness in downstream applications.

%
% To improve visual document retrieval, recent studies~\cite{faysse2025colpali, yu2025visrag, tanaka2025vdocrag} usually employ a visual retriever that directly encodes document screenshots.
% Although direct page-image encoding better preserves layout-dependent evidence than OCR-based serialization, most visual retrievers are still supervised only at the page level using contrastive losses such as InfoNCE~\cite{oord2018representation}, which indicate whether a query matches a page but do not reveal which local evidence justifies the match, leaving models struggling to capture critical fine-grained visual cues. 
% To mitigate this issue, some methods introduce query-relevant local evidence signals, such as region-level matching~\cite{faysse2025colpali, cui2025attention} or localized region descriptions~\cite{yang2026realign, abdallah2026argus}, to guide models toward important page areas. 
% While effective, text-based evidence descriptions project localized visual evidence into the textual semantic space, reducing visually grounded semantics.
To improve visual document retrieval, recent studies~\cite{faysse2025colpali,tanaka2025vdocrag,yu2025visrag} leverage the visual understanding capabilities of VLMs to encode queries and document screenshots into dense representations and optimize query-document alignment in a shared embedding space through contrastive training~\cite{oord2018representation}. However, most existing visual retrievers are still supervised only at the page level through contrastive objectives. Such supervision provides a coarse-grained relevance signal but does not specify the supporting visual evidence within document pages, making it challenging for models to identify critical fine-grained cues~\cite{yang2026realign}.
To address this limitation, recent studies introduce query-relevant fine-grained evidence signals for visual document retrieval, including region-level matching~\cite{cui2025attention,faysse2025colpali} and localized region-based textual descriptions~\cite{abdallah2026argus,yang2026realign}, to guide models toward informative page regions. Nevertheless, both approaches have inherent limitations in representing query-relevant evidence. Textual descriptions provide high-level semantic interpretations but may fail to faithfully capture complex visual structures, such as charts and spatial layouts. In contrast, localized visual regions preserve fine-grained visual grounding but often lack the global context and document-level semantics necessary for accurately modeling page relevance.

Recent advances in multimodal latent reasoning~\cite{pham2025multimodal,li2025latent,viveiros2026lantern} investigate continuous latent representations as an alternative to natural language for modeling intermediate reasoning. By encoding reasoning into latent visual states, enabling multimodal interaction in a shared latent space, or interleaving language with continuous visual representations, these approaches have achieved strong performance across visual understanding, reasoning, and generation tasks. This emerging paradigm also opens up a promising direction for visual document retrieval: latent representations may provide a more flexible and expressive mechanism for modeling query-relevant evidence, thereby facilitating finer-grained alignment between textual queries and visual documents~\cite{liuuniversal}.

In this paper, we propose \method{}, a latent concept representation learning framework for visual document retrieval. \method{} represents query-relevant evidence as continuous, query-conditioned latent concepts, providing fine-grained query-document alignment signals to narrow the semantic gap between textual queries and visual documents. Specifically, given a query-document pair, \method{} first leverages a strong vision-language model to localize query-relevant regions on the positive page and maps these regions onto the retriever's visual token grid to estimate the required capacity of latent concepts for evidence representation. It then dynamically allocates a latent concept sequence with the corresponding length and enables the retriever to generate latent concept representations conditioned on query and its related document. To leverage these representations as training supervision, \method{} optimizes the retriever such that the ranking distribution induced by latent concepts recovers the original query-driven document ranking preference.

Experimental results on diverse visual document retrieval benchmarks demonstrate the effectiveness of \method{}, achieving significant improvements over existing retrieval approaches. This improvement can be attributed to the learned latent concepts, which provide more effective and fine-grained supervision beyond standard page-level matching, enabling the retriever to learn more tailored visual document representations, suggesting that adaptive latent concepts provide complementary relevance supervision for visual document retrieval. Further analyses reveal that the learned latent concepts occupy an intermediate space between queries and visual documents, making them more effective for cross-modal alignment than textual or visual concept proxies. Moreover, the adaptive capacity analysis shows that query-relevant evidence exhibits substantial variation in scale rather than following a fixed-size pattern. Dynamically allocating latent concept capacity according to evidence coverage therefore provides a more tailored and sufficient concept of query-related evidence in visual documents.
\section{Related Work}
Visual document retrieval aims to retrieve document pages relevant to a query from visually rich collections~\cite{marinai2011digital,doermann1998indexing,alaei2016brief}. Early approaches typically rely on OCR to convert page images into textual representations~\cite{mitra2000information, barboule2025survey, taghva1994results, ding2026deep}, followed by retrieval using sparse methods such as BM25~\cite{robertson2009probabilistic}, dense retrievers~\cite{bge_embedding, sturua2024jina}, or LLM-based embeddings~\cite{lee2025nv, wang2024improving, yao2025expandr}. However, OCR-based pipelines are susceptible to recognition errors and may discard essential visual information, including spatial layouts, chart structures, and non-textual elements. To mitigate information loss introduced by document parsing, recent studies~\cite{koukounas2024jinaclipv2, yu2025visrag, yang2026realign, peng2026memory} have explored direct retrieval from page images, focusing on adapting VLMs to jointly model query and visual document representations. For example, DSE~\cite{ma2024unifying} learns unified representations from document screenshots and employs contrastive learning to optimize VLMs for retrieval-oriented representation learning. VisRAG~\cite{yu2025visrag} constructs additional query-document pairs to improve retrieval training. VDocRetriever~\cite{tanaka2025vdocrag} further leverages LVLM pre-training to transfer their strong visual understanding and generative capabilities to visual document retrieval. However, most existing approaches still perform retrieval through global matching between queries and holistic page representations, limiting their ability to identify fine-grained query-relevant evidence within complex visual documents.

To further enhance visual document representations, recent studies~\cite{lin2023fine, lin2024preflmr} have explored fine-grained modeling strategies for visual documents. ColPali~\cite{faysse2025colpali} improves query-page matching through multi-vector interactions that capture localized page-level semantics. Other approaches~\cite{tanaka2025vdocrag, yuan2023vile, li2025docmmir} explicitly verbalize visual evidence and leverage the resulting textual descriptions as auxiliary signals to enhance retrieval. ReAlign~\cite{yang2026realign} further exploits the reasoning capabilities of existing VLMs to identify query-relevant regions and generate descriptions for the cropped visual evidence. These region-level descriptions provide additional supervision for retriever training. Despite the effectiveness of local evidence modeling, most existing visual document retrievers still rely on fixed-granularity alignment between queries and page representations, leaving the potential of reasoning-guided and dynamically adaptive supervision largely unexplored.

To enable finer-grained visual reasoning in vision-language models~\cite{sun2025visrag, xiong2026lang2act}, recent studies have explored constructing intermediate reasoning representations in continuous latent spaces~\cite{yu2026latent, hao2024training, chen2025reasoning}. LVR~\cite{li2025latent} enables models to learn latent visual states that preserve query-relevant visual information. MCOUT~\cite{pham2025multimodal} transfers multimodal reasoning from natural-language chain-of-thought into a unified latent space. LanteRn~\cite{viveiros2026lantern} further interleaves language reasoning with continuous latent visual representations, demonstrating that latent states can encode visual reasoning signals beyond explicit textual rationales. However, existing latent reasoning approaches typically rely on fixed-length latent representations and are not designed to capture the query-dependent evidence granularity required for visual document retrieval~\cite{yang2026machine, chen2025reasoning, jiang2026univlr}. In contrast, \method{} introduces retrieval-oriented latent concepts for visual document retrieval, dynamically adapting their capacity to query-relevant evidence coverage while shaping their representations through query-specific ranking signals.
\section{Methodology}
\begin{figure*}[t]
    \centering
    \includegraphics[width=1.0\linewidth]{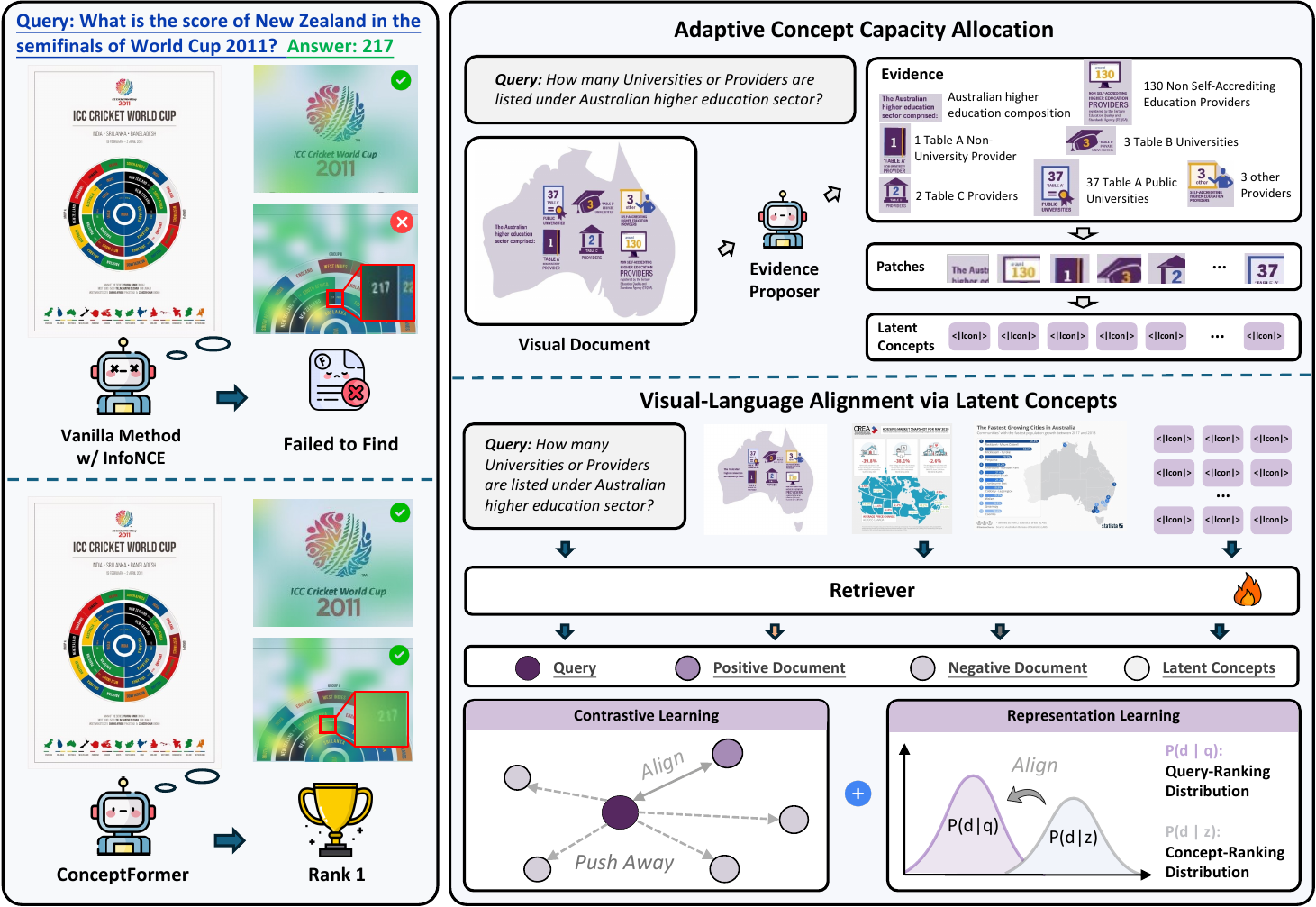}
    \caption{Overview of \method{}. The left panel illustrates a case where the standard InfoNCE method fails to find fine-grained local evidence, whereas \method{} retrieves the relevant page. The right panel shows how \method{} allocates adaptive latent concepts from query-related regions and learns them with retrieval-guided representation learning.}
    \label{fig:method}
\end{figure*}
In this section, we first introduce the standard formulation of visual document retrieval and its contrastive training objective in Section~\ref{sec:preliminary}. We then describe \method{} in Section~\ref{sec:\method{}}, which augments the retriever with a dynamic latent concept space for modeling query-relevant information within visually rich pages. 

\subsection{Preliminaries of Visual Document Retrieval}
\label{sec:preliminary}
Visual document pages organize heterogeneous information, including textual content, layout structures, tables, charts, and visual symbols, in pure image form. We formulate visual document retrieval as a cross-modal matching problem over a collection of candidate pages $\mathcal{D}=\{d_1,\ldots,d_N\}$. For each natural language query $q_i$, the retriever compares $q_i$ with every candidate page $d_j\in\mathcal{D}$ in a shared semantic space by encoding them into dense representations:
\begin{equation}
    \mathbf{q}_i = f_q(q_i), \qquad \mathbf{d}_j = f_d(d_j),
\end{equation}
where $f_q(\cdot)$ and $f_d(\cdot)$ denote the query encoder and the document encoder, respectively. The relevance between $q_i$ and $d_j$ is then measured by a similarity function:
\begin{equation}
s(q_i,d_j) =\operatorname{sim}(\mathbf{q}_i,\mathbf{d}_j),
\end{equation}
where $\operatorname{sim}(\cdot,\cdot)$ is typically instantiated as cosine similarity~\cite{radford2021learning,yu2025visrag}. During retrieval, all candidate pages are ranked according to their similarity scores with the query.

The retriever is trained with contrastive learning, considering a mini-batch of $B$ query-page pairs $\{(q_i,d_i^+)\}_{i=1}^{B}$, where $d_i^+$ is the positive page associated with $q_i$. For each query, the remaining pages in the mini-batch naturally form negative candidates, optionally together with additional sampled negatives. We denote the resulting candidate set for $q_i$ as:
\begin{equation}
\widetilde{\mathcal{D}}_i = \{d_i^+\} \cup \mathcal{D}_i^- ,
\end{equation}
where $\mathcal{D}_i^-$ contains the in-batch negative pages for $q_i$. Based on the similarity scores over $\widetilde{\mathcal{D}}_i$, the model defines a ranking distribution:
\begin{equation}
\label{eq:distribution}
P(d \mid q_i,\widetilde{\mathcal{D}}_i)=\frac{\exp(s(q_i,d)/\tau)}{\sum_{d' \in \widetilde{\mathcal{D}}_i}\exp(s(q_i,d')/\tau)},
\end{equation}
where $\tau$ is the temperature coefficient, controlling the sharpness of the ranking distribution. The standard retrieval objective maximizes the probability assigned to the positive page:
\begin{equation}
\mathcal{L}_{\mathrm{cons}}
=
-\frac{1}{B}
\sum_{i=1}^{B}
\log
P(d_i^+ \mid q_i,\widetilde{\mathcal{D}}_i).
\end{equation}
This objective learns a shared embedding space in which each query is close to its relevant page and separated from irrelevant pages. \method{} builds on this standard retrieval formulation by introducing adaptive latent concepts to model query-conditioned relevance signals within visually rich pages.

\subsection{Bringing Visual-Language Modality Gap Using Latent Concepts}
\label{sec:\method{}}

As shown in Figure~\ref{fig:method}, \method{} augments a standard visual document retriever with a training-time latent concept space that explicitly connects localized visual grounding with query-level semantic relevance. Different from existing visual retrievers~\cite{radford2021learning,yu2025visrag}, \method{} introduces latent concepts as an intermediate representation: a continuous, query-conditioned relevance space grounded in page regions and capturing the semantic evidence required for document ranking. Specifically, we first leverage Matryoshka Representations to adaptively estimate the concept capacity required for each query-document pair, enabling the model to allocate different levels of semantic abstraction according to retrieval difficulty. The resulting latent concepts provide structured semantic anchors that guide region-level visual representations toward query-relevant semantics, thereby reducing the visual-language modality gap and improving fine-grained visual document retrieval.

\noindent\paragraph{Concept Learning via Matryoshka Representations.}
To adaptively allocate the latent concept capacity for each query-document pair, \method{} employs a powerful vision-language model $\mathcal{M}$ as a training-time evidence proposer. Given a training pair $(q_i,d_i^+)$, where $q_i$ denotes the query and $d_i^+$ represents the positive document page, $\mathcal{M}$ is prompted to identify query-relevant visual evidence by jointly reasoning over the textual query and the page image:
\begin{equation}
R_i = \mathcal{M}(\text{Instruct}_{\text{evidence}}, q_i, d_i^+),
\end{equation}
where $\text{Instruct}_{\text{evidence}}$ specifies the evidence extraction instruction, and $R_i$ denotes the generated evidence description. We further parse $R_i$ into structured visual evidence annotations:
\begin{equation}
\label{eq:label}
\{b_{i,n}\}_{n=1}^{N_i} = \text{Parse} (R_i),
\end{equation}
where $N_i$ indicates the number of evidence regions identified for the query-document pair. $b_{i,n}=(x_{i,n}^{1},y_{i,n}^{1},x_{i,n}^{2},y_{i,n}^{2})$ denotes the bounding box of the $n$-th evidence region, with $(x_{i,n}^{1},y_{i,n}^{1})$ and $(x_{i,n}^{2},y_{i,n}^{2})$ representing the top-left and bottom-right coordinates, respectively. 
These evidence annotations provide reasoning-guided spatial grounding, following the reasoning-crop-observation paradigm, where textual reasoning is coupled with localized visual regions for fine-grained evidence verification. Such structured evidence enables \method{} to estimate the semantic complexity of each query-document pair and adaptively determine its required latent concept capacity. Specifically, samples involving richer and more diverse visual evidence are allocated larger concept capacities, whereas simpler cases are represented with fewer latent concepts.

% To adapt the capacity of the latent concept space to the evidence scale of each query-document pair, \method{} leverages a strong vision-language model $\mathcal{M}$ as an evidence proposer during training. Given a training pair $(q_i,d_i^+)$, $\mathcal{M}$ takes the query $q_i$ and the positive page image $d_i^+$ as input, and produces a set of query-related evidence proposals:
% \begin{equation}
% \mathcal{E}_i=\{(b_{i,n}, r_{i,n})\}_{n=1}^{N_i},
% \end{equation}
% where $b_{i,n}=(x_{i,n}^{1}, y_{i,n}^{1}, x_{i,n}^{2}, y_{i,n}^{2})$ denotes the bounding box of the $n$-th evidence region with $(x_{i,n}^{1}, y_{i,n}^{1})$ and $(x_{i,n}^{2}, y_{i,n}^{2})$ corresponding to its top-left and bottom-right corners, respectively, which provides the spatial region that can be mapped onto the retriever's visual token grid, $r_{i,n}$ denotes the corresponding textual cue, which records the semantic interpretation of the localized evidence, and $N_i$ is the number of evidence regions identified for this query-document pair. \method{} uses the visual coverage of these query-relevant regions to estimate how much latent concept capacity should be assigned to the current sample.

Since different retrievers may partition the same page image into different visual token grids, \method{} determines the latent concept capacity based on the tokenization strategy of the retrieval model $\pi_\theta$. Specifically, given an evidence region $b_{i,n}$, we project its coordinates onto the visual patch grid produced by $\pi_\theta$ and identify the patches that overlap with this region:
\begin{equation}
\label{eq:index}
\mathcal{I}_{i,n}
=
\{m \mid P_m(d_i^+) \cap b_{i,n} \neq \emptyset\},
\end{equation}
where $P_m(d_i^+)$ denotes the image region corresponding to the $m$-th visual patch of page $d_i^+$, and $\mathcal{I}_{i,n}$ contains the indices of visual patches covered by the $n$-th evidence region. We define the number of covered patches:
\begin{equation}
t_{i,n}=|\mathcal{I}_{i,n}|,
\end{equation}
as the local concept scale, which reflects the visual capacity required to represent the grounded evidence and align it with query semantics. For a query-document pair with multiple evidence regions, \method{} aggregates their local concept scales to derive the adaptive latent concept length:
\begin{equation}
T_i=\sum_{n=1}^{N_i}t_{i,n}.
\end{equation}
Samples with more localized evidence receive shorter latent concept segments, while samples involving large charts, maps, tables, or cross-region relations are assigned larger concept capacity. Given $T_i$, \method{} allocates a sequence of latent concept tokens:
\begin{equation}
\ell_i=[\ell_{i,1},\ell_{i,2},\ldots,\ell_{i,T_i}],
\end{equation}
where each $\ell_{i,k}$ is instantiated as the same special token \texttt{<|lcon|>}. 
% which do not carry natural-language content at the input level. Instead, they serve as latent concept slots for the current query-document pair, whose hidden states are later formed under the query-page context and optimized through latent concept alignment.

\noindent\paragraph{Visual-Language Alignment via Latent Concepts.}
After determining the adaptive latent concept length $T_i$, \method{} introduces a sequence of latent concept tokens $\ell_i=\{\ell_{i,1},...,\ell_{i,T_i}\}$ into the retriever context and derives continuous concept states from the retrieval model itself. Specifically, for the $k$-th latent concept token, the retriever $\pi_\theta$ produces a hidden representation conditioned on the query $q_i$, the positive page $d_i^+$, the concept initialization token \texttt{<|lcon\_start|>}, and the preceding latent concept tokens:
\begin{equation}
\mathbf{c}_{i,k}
=\pi_\theta(q_i,d_i^+,\texttt{<|lcon\_start|>},\ell_{i,1:k}),
\end{equation}
where $\ell_{i,1:k}$ denotes the latent concept token prefix up to the $k$-th position. The resulting hidden states 
$\mathbf{c}_{i,1:T_i}$ form a query-conditioned latent concept sequence.

We aggregate these states to obtain a compact latent concept representation:
\begin{equation}
\hat{\mathbf{c}}_i=\operatorname{MeanPool}(\mathbf{c}_{i,1},...,\mathbf{c}_{i,T_i}),
\end{equation}
where $\hat{\mathbf{c}}_i$ serves as an intermediate relevance representation learned within the retriever. Different from raw visual patch features or textual descriptions, it encodes query-conditioned semantic relevance grounded by localized evidence regions through the dynamically allocated concept capacity.

Following the ranking distribution in Eq.~\ref{eq:distribution}, the query representation induces a retrieval preference distribution over candidates:
\begin{equation}
P(d|q_i,\tilde{\mathcal{D}}_i).
\end{equation}
Similarly, replacing the query representation with the latent concept  $\hat{\mathbf{c}}_i$, we obtain the concept-induced ranking distribution:
\begin{equation}
P(d|\hat{\mathbf{c}}_i,\tilde{\mathcal{D}}_i).
\end{equation}
The query-induced distribution acts as a teacher that provides ranking behavior supervision for the latent concept representation. Specifically, \method{} minimizes their discrepancy using KL divergence:
\begin{equation}
\mathcal{L}_{\mathrm{align}}
=
\frac{1}{B}
\sum_{i=1}^{B}
\sum_{d\in\tilde{\mathcal{D}}_i}
P(d|q_i,\tilde{\mathcal{D}}_i)
\log
\frac{
P(d|q_i,\tilde{\mathcal{D}}_i)
}{
P(d|\hat{\mathbf{c}}_i,\tilde{\mathcal{D}}_i)
}.
\end{equation}
This objective encourages the latent concepts to preserve the retrieval behavior of the original query while encoding the relevance factors required for document ranking. Consequently, the concept-aware supervision is propagated through $\pi_\theta$, enabling the retriever to capture fine-grained visual evidence that contributes to query-page relevance.
Finally, the overall training objective is:
\begin{equation}
\label{eq:loss}
\mathcal{L}
=
\mathcal{L}_{\mathrm{cons}}
+
\lambda\mathcal{L}_{\mathrm{align}},
\end{equation}
where $\mathcal{L}_{\mathrm{cons}}$ optimizes the original page-level query-document discrimination, and $\mathcal{L}_{\mathrm{align}}$ aligns the ranking behavior induced by latent concepts with that of the query representation. Their joint optimization preserves the standard retrieval objective while introducing an adaptive concept-level relevance pathway that connects localized visual grounding with query semantics. During inference, \method{} follows the original visual document retrieval pipeline without requiring latent concept generation.

\section{Experimental Methodology}
In this section, we describe datasets, evaluation metrics, baseline methods, and implementation details.

\textbf{Datasets.}
Following the experimental settings of VDocRAG~\cite{tanaka2025vdocrag}, we construct the training and evaluation datasets for visual document retrieval. Specifically, the training set contains approximately 38,000 query-document pairs collected from DocVQA~\cite{mathew2021docvqa}, InfoVQA~\cite{mathew2022infographicvqa}, VisualMRC~\cite{tanaka2021visualmrc}, OpenWikiTable~\cite{kweon2023open}, DUDE~\cite{van2023document}, and MHDocVQA~\cite{tanaka2025vdocrag}. These datasets cover a broad range of visually rich document types, including industrial documents, infographics, web pages, tables, and open-domain document pages.
For evaluation, we benchmark retrieval models on six visual document retrieval datasets. InfoVQA~\cite{mathew2022infographicvqa} serves as the in-domain evaluation benchmark, while ChartQA~\cite{masry2022chartqa}, SlideVQA~\cite{tanaka2023slidevqa}, TQA~\cite{kembhavi2017you}, OWID Charts~\cite{OurWorldInData}, and Wikimedia Maps~\cite{gunther2025jina} are used for out-of-domain evaluation. These datasets cover diverse visual document formats, including charts, slides, tables, statistical graphics, and maps. Detailed data statistics are provided in Appendix~\ref{app:datasets}.

\textbf{Evaluation Metrics.}
Following previous studies~\cite{tanaka2025vdocrag, yang2026realign}, we adopt Recall@10 and NDCG@10 as evaluation metrics. Statistic significances are tested by permutation test with $P< 0.05$.

\textbf{Baselines.}
We compare \method{} with two categories of retrieval approaches: OCR-based text retrievers and image-based visual retrievers. More details about baseline models, checkpoints, and input configurations is provided in Appendix~\ref{app:baselines}.

For OCR-based text retrievers, we first extract textual information from document images using PaddleOCR~\cite{cui2025paddleocr}, and then perform retrieval based on the extracted text. This category includes BM25~\cite{robertson2009probabilistic}, a classical sparse retrieval method based on lexical matching; BGE~\cite{bge_embedding}, a representative dense text retriever; and E5-Mistral-7B-Instruct~\cite{wang2024improving} and NV-Embed-v2~\cite{lee2025nv}, two LLM-based embedding models with strong text representation capabilities.

For visual retrievers, retrieval is performed directly on document page images without relying on OCR extraction. We include CLIP~\cite{radford2021learning} as a general image-text dual encoder; DSE~\cite{ma2024unifying} as a screenshot-based document retriever; E5-V~\cite{jiang2024e5} and VLM2Vec~\cite{jiang2024vlm2vec} as multimodal embedding models built upon vision-language models; Jina-CLIP-v2~\cite{koukounas2024jinaclipv2} as a strong image-text retrieval model; and VDocRetriever~\cite{tanaka2025vdocrag} and VisRAG-Ret~\cite{yu2025visrag} as retrieval models specifically designed for visual document RAG. 

\textbf{Implementation Details.}
During training, we employ Qwen3.6-Plus~\cite{qwen36plus} is only using in the training process to determine the adaptive capacity of latent concepts. The retriever backbone is initialized with Phi3V-4B~\cite{abdin2024phi} and Qwen2.5-VL-7B-Instruct~\cite{bai2025qwen2}, enabling evaluation across different vision-language encoder architectures. 

All variants of \method{} are trained for 3 epochs using the AdamW optimizer~\cite{loshchilov2017decoupled} with an effective batch size of $B=128$. The weight of the latent concept representation loss $\mathcal{L}_{\mathrm{repr}}$ is set to $\lambda=0.2$. Experiments are conducted on four A100 GPUs with 40GB memory each. Unless otherwise specified, all models share the same training and inference configurations for fair comparison. Additional implementation details are provided in Appendix~\ref{app:implementation}.
\begin{table*}[t]
\centering
\caption{Overall performance of \method{}. We report Recall@10 and NDCG@10 as evaluation metrics. ${\dagger}$, ${\ddagger}$, and ${\S}$ indicate statistically significant improvements over NV-Embed$^{\dagger}$, VisRAG-Ret$^{\ddagger}$, and VDocRetriever$^{\S}$, respectively.}

\resizebox{0.98\textwidth}{!}{
\label{tab:overall}
\begin{tabular}{lcccccccccccccc}
\toprule
\multirow{2}{*}{Method}
& \multicolumn{2}{c}{InfoVQA}
& \multicolumn{2}{c}{ChartQA}
& \multicolumn{2}{c}{SlideVQA}
& \multicolumn{2}{c}{TQA}
& \multicolumn{2}{c}{OWID Charts}
& \multicolumn{2}{c}{Wikimedia Maps}
& \multicolumn{2}{c}{Average} \\
& R@10 & N@10
& R@10 & N@10
& R@10 & N@10
& R@10 & N@10
& R@10 & N@10
& R@10 & N@10
& R@10 & N@10 \\
\hline
\rowcolor{black!5}\multicolumn{15}{l}{\emph{Text-based retrievers}}\\
BM25 & 81.77 & 68.91 & 58.67 & 48.88 & 81.43 & 73.55 & 24.00 & 12.73 & 98.47 & 88.27 & 5.93 & 3.26 & 58.38 & 49.27 \\
BGE-1.5 & 79.87 & 64.27 & 82.00 & 70.83 & 81.36 & 70.47 & 43.40 & 22.61 & \textbf{99.24} & 91.92 & 9.45 & 5.95 & 65.89 & 54.34 \\
E5-Mistral & 54.48 & 40.31 & 55.33 & 47.78 & 62.08 & 50.42 & 32.60 & 17.45 & \textbf{99.24} & 89.23 & 6.59 & 4.08 & 51.72 & 41.55 \\
NV-Embed-v2 & 88.17 & 73.66 & 88.67 & 80.91 & 88.25 & 79.40 & 61.80 & 34.77 & \textbf{99.24} & 94.79 & 15.82 & 9.92 & 73.66 & 62.24 \\
\hline
\rowcolor{black!5}\multicolumn{15}{l}{\emph{Multimodal retrievers}}\\
CLIP & 58.21 & 42.52 & 61.33 & 50.27 & 50.46 & 36.98 & 44.40 & 24.06 & 98.47 & 85.07 & 34.29 & 24.22 & 57.86 & 43.85 \\
Jina-CLIP-v2 & 53.91 & 37.18 & 66.67 & 53.14 & 57.70 & 45.30 & 33.10 & 17.35 & 96.95 & 81.86 & 35.16 & 24.80 & 57.25 & 43.27 \\
DSE & 84.92 & 71.36 & 89.33 & 82.90 & 85.75 & 76.15 & 70.40 & 39.46 & \textbf{99.24} & 91.42 & 39.12 & 29.01 & 78.13 & 65.05 \\
E5-V & 61.74 & 43.84 & 85.33 & 75.71 & 61.03 & 48.15 & 53.60 & 28.14 & \textbf{99.24} & 93.14 & 27.91 & 20.13 & 64.81 & 51.52 \\
VLM2Vec & 64.22 & 45.00 & 63.33 & 48.72 & 51.56 & 38.31 & 34.90 & 18.43 & 88.55 & 67.78 & 30.55 & 19.43 & 55.52 & 39.61 \\
ColPali & 78.05 & 61.58 & 92.67 & 82.33 & 84.41 & 74.86 & 71.60 & 40.99 & 98.47 & 87.19 & 30.55 & 22.73 & 75.96 & 61.61 \\
VDocRetriever & 83.21 & 68.54 & 95.33 & 86.79 & 87.94 & 78.96 & 67.50 & 37.19 & \textbf{99.24} & 91.71 & 33.85 & 22.59 & 77.84 & 64.30 \\
VisRAG-Ret & 86.36 & 69.16 & 92.00 & 85.14 & 81.91 & 71.92 & 58.30 & 32.18 & \textbf{99.24} & 93.46 & 44.84 & 29.04 & 77.11 & 63.48 \\
\hline
\rowcolor{black!5}\multicolumn{15}{l}{\emph{Ours}}\\
\method{} (Phi3V) & 91.70\rlap{$^{\dagger \ddagger \S}$} & 76.41\rlap{$^{\dagger \ddagger \S}$} & 95.33\rlap{$^{\dagger}$} & 88.66\rlap{$^{\dagger \ddagger}$} & 88.73\rlap{$^{\ddagger}$} & 78.56\rlap{$^{\ddagger}$} & 68.80\rlap{$^{\dagger \S}$} & 39.53\rlap{$^{\dagger \ddagger \S}$} & \textbf{99.24} & 92.98 & 36.70\rlap{$^{\dagger \ddagger}$} & 25.89\rlap{$^{\dagger \ddagger \S}$} & 80.08\rlap{$^{\dagger \ddagger \S}$} & 67.00\rlap{$^{\dagger \ddagger \S}$} \\
\method{} (Qwen) & \textbf{93.03}\rlap{$^{\dagger \ddagger \S}$} & \textbf{79.23}\rlap{$^{\dagger \ddagger \S}$} & \textbf{98.67}\rlap{$^{\dagger \ddagger \S}$} & \textbf{95.79}\rlap{$^{\dagger \ddagger \S}$} & \textbf{90.88}\rlap{$^{\dagger \ddagger \S}$} & \textbf{82.41}\rlap{$^{\dagger \ddagger \S}$} & \textbf{72.10}\rlap{$^{\dagger \ddagger \S}$} & \textbf{41.30}\rlap{$^{\dagger \ddagger \S}$} & \textbf{99.24} & \textbf{95.39}\rlap{$^{\ddagger \S}$} & \textbf{76.04}\rlap{$^{\dagger \ddagger \S}$} & \textbf{61.69}\rlap{$^{\dagger \ddagger \S}$} & \textbf{88.33}\rlap{$^{\dagger \ddagger \S}$} & \textbf{75.97}\rlap{$^{\dagger \ddagger \S}$} \\
\bottomrule
\end{tabular}}
\end{table*}
\section{Evaluation Results}
In this section, we evaluate the retrieval effectiveness of \method{}, analyze the impact of its key components, and investigate the learned latent concept space and adaptive concept capacity allocation. Finally, we conduct case studies.

\subsection{Overall Performance}
As shown in Table \ref{tab:overall}, we present the retrieval performance of \method{} and baseline methods across six visual document retrieval benchmarks.
Overall, \method{} achieves the best average performance among all compared approaches, yielding relative improvements of 16.7\% and 22.1\% in average NDCG@10 over the strongest visual retriever and OCR-based text retriever, respectively. These results demonstrate that adaptive latent concept learning provides consistent benefits beyond conventional page-level representation learning. Notably, on the OWID Charts~\cite{OurWorldInData} dataset, where several strong baselines already achieve competitive retrieval coverage, \method{} attains the highest ranking performance, indicating its ability to more effectively prioritize highly relevant pages among top-ranked results.

Compared with OCR-based retrievers, \method{} achieves substantial improvements by directly encoding visual document pages, preserving visual structures such as layouts, charts, and other non-textual cues that may be lost during text extraction. Compared with existing visual retrievers, \method{} further introduces a latent concept space that improves semantic alignment between queries and visual documents, thereby enhancing page-level retrieval beyond global page representations. On Wikimedia Maps~\cite{gunther2025jina}, \method{} achieves over 2$\times$ the NDCG@10 of the strongest prior visual retriever, suggesting that adaptive latent concepts are particularly beneficial for scenarios where relevant evidence is spatially distributed and difficult to capture using a single global page representation.

Furthermore, \method{} demonstrates consistent improvements across different VLM backbones. The Phi3V-based~\cite{abdin2024phi} variant already outperforms the strongest existing visual retriever by approximately 4\% on average, showing that the proposed framework remains effective even with a relatively weaker backbone. Using the stronger Qwen backbone~\cite{bai2025qwen2}, \method{} achieves an additional improvement of over 10\% compared with its Phi3V counterpart. These results suggest that \method{} is broadly compatible with different VLM architectures, and its improvements stem not only from backbone scaling but also from latent concept guided representation learning.

\begin{table*}[t]
\centering
\caption{Ablation study of latent concept representation learning in \method{}. ${\dagger}$ and ${\ddagger}$ indicate statistically significant improvements over \method{} (Text)$^{\dagger}$, \method{} w/o KL Loss$^{\ddagger}$, respectively.}

\resizebox{1.0\textwidth}{!}{
\label{tab:ablation_align}
\begin{tabular}{lcccccccccccccc}
\toprule
\multirow{2}{*}{Models}
& \multicolumn{2}{c}{InfoVQA}
& \multicolumn{2}{c}{ChartQA}
& \multicolumn{2}{c}{SlideVQA}
& \multicolumn{2}{c}{TQA}
& \multicolumn{2}{c}{OWID Charts}
& \multicolumn{2}{c}{Wikimedia Maps}
& \multicolumn{2}{c}{Average} \\
& R@10 & N@10
& R@10 & N@10
& R@10 & N@10
& R@10 & N@10
& R@10 & N@10
& R@10 & N@10
& R@10 & N@10 \\
\midrule
\method{} (Text) & 90.27 & 77.04 & \textbf{100.00} & 95.61 & \textbf{91.12} & 81.43 & 71.50 & 41.14 & 99.24 & 94.82 & 74.73 & 59.08 & 87.81 & 74.85 \\
\method{} (Vision) & 89.22 & 75.91 & 98.67 & 92.78 & 89.04 & 79.37 & 68.60 & 38.64 & 99.24 & 95.37 & 70.11 & 54.79 & 85.81 & 72.81 \\
Query-Concept Matching & 92.37 & 79.06 & 99.33 & 94.87 & 90.22 & 80.63 & 72.00 & \textbf{41.61} & 99.24 & 94.67 & 75.60 & 60.04 & 88.13 & 75.15 \\
\method{} & \textbf{93.03}\rlap{$^{\dagger\ddagger}$} & \textbf{79.23}\rlap{$^{\dagger\ddagger}$} & 98.67 & \textbf{95.79} & 90.88 & \textbf{82.41}\rlap{$^{\dagger\ddagger}$} & \textbf{72.10} & 41.30 & 99.24 & 95.39 & \textbf{76.04}\rlap{$^{\dagger\ddagger}$} & \textbf{61.69}\rlap{$^{\dagger\ddagger}$} & \textbf{88.33}\rlap{$^{\dagger\ddagger}$} & \textbf{75.97}\rlap{$^{\dagger\ddagger}$} \\
$\quad$ w/o Mean Pooling & 92.56 & 78.95 & 99.33 & 94.78 & 90.48 & 81.65 & 71.80 & 41.21 & 99.24 & 95.29 & 74.51 & 60.69 & 87.99 & 75.43 \\
$\quad$ w/o KL Loss & 90.55 & 76.86 & \textbf{100.00} & 95.00 & 89.50 & 79.01 & 69.80 & 40.50 & 99.24 & \textbf{96.03} & 74.07 & 57.91 & 87.19 & 74.22 \\
% $\quad$ + MSE Loss & 91.70 & 77.65 & 98.67 & 95.31 & 90.83 & 80.91 & \textbf{72.40} & \textbf{42.58} & 99.24 & 95.01 & 71.87 & 58.06 & 87.45 & 74.92 \\
% $\quad$ + Cosine Loss & 91.89 & \textbf{79.25} & 99.33 & 94.91 & 90.24 & 80.32 & 70.90 & 40.95 & 99.24 & 95.95 & 71.43 & 59.05 & 87.17 & 75.07 \\
\bottomrule
\end{tabular}}
\end{table*}
\begin{table}[t]
\small
\centering
\caption{Sensitivity to the weight $\lambda$ of alignment loss $\mathcal{L}_{\mathrm{align}}$ (Eq.~\ref{eq:loss}). All models are implemented using Qwen2.5-vl.}
\label{tab:ablation_lambda}

\resizebox{\linewidth}{!}{\begin{tabular}{lccccccc}
\toprule
$\lambda$
& I-VQA
& C-tQA
& S-VQA
& TQA
& O-Charts
& W-Maps
& Average \\
\hline
0.0   & 76.86 & 95.00 & 79.01 & 40.50 & 96.03 & 57.91 & 74.22 \\
0.1 & 74.33 & 93.11 & 79.24 & 38.54 & 95.52 & 58.37 & 73.18 \\
0.2 & \textbf{79.23} & \textbf{95.79} & \textbf{82.41} & 41.30 & 95.39 & \textbf{61.69} & \textbf{75.97} \\
0.3 & 77.34 & 93.72 & 79.72 & 40.00 & 96.13 & 57.79 & 74.12 \\
0.4 & 76.80 & 93.10 & 79.83 & 41.31 & 95.85 & 57.50 & 74.06 \\
0.5 & 78.89 & 93.72 & 79.54 & \textbf{41.48} & \textbf{96.23} & 56.36 & 74.37 \\
% \hline
% \rowcolor{black!5}\multicolumn{8}{l}{\emph{Phi3V w/ Pre-training}}\\
% 0.0 & 75.94 & 86.87 & 78.82 & 39.36 & 92.04 & 24.42 & 66.24 \\
% 0.1 & 76.30 & \textbf{88.76} & 78.96 & 39.38 & 92.34 & 25.59 & 66.89 \\
% 0.2 & 76.41 & 88.66 & 78.56 & 39.53 & 92.98 & 25.89 & 67.00 \\
% 0.3 & 76.26 & 88.31 & 79.10 & 39.37 & 92.72 & 25.59 & 66.89 \\
% 0.4 & \textbf{76.51} & \textbf{88.76} & 79.03 & 39.26 & 92.59 & 26.24 & 67.07 \\
% 0.5 & 76.38 & 87.97 & \textbf{79.11} & \textbf{39.80} & \textbf{93.08} & \textbf{26.43} & \textbf{67.13} \\
\bottomrule
\end{tabular}}
\end{table}

\begin{figure}[t]
    \centering
    \begin{subfigure}[t]{0.48\linewidth}
        \centering
        \includegraphics[width=\linewidth]{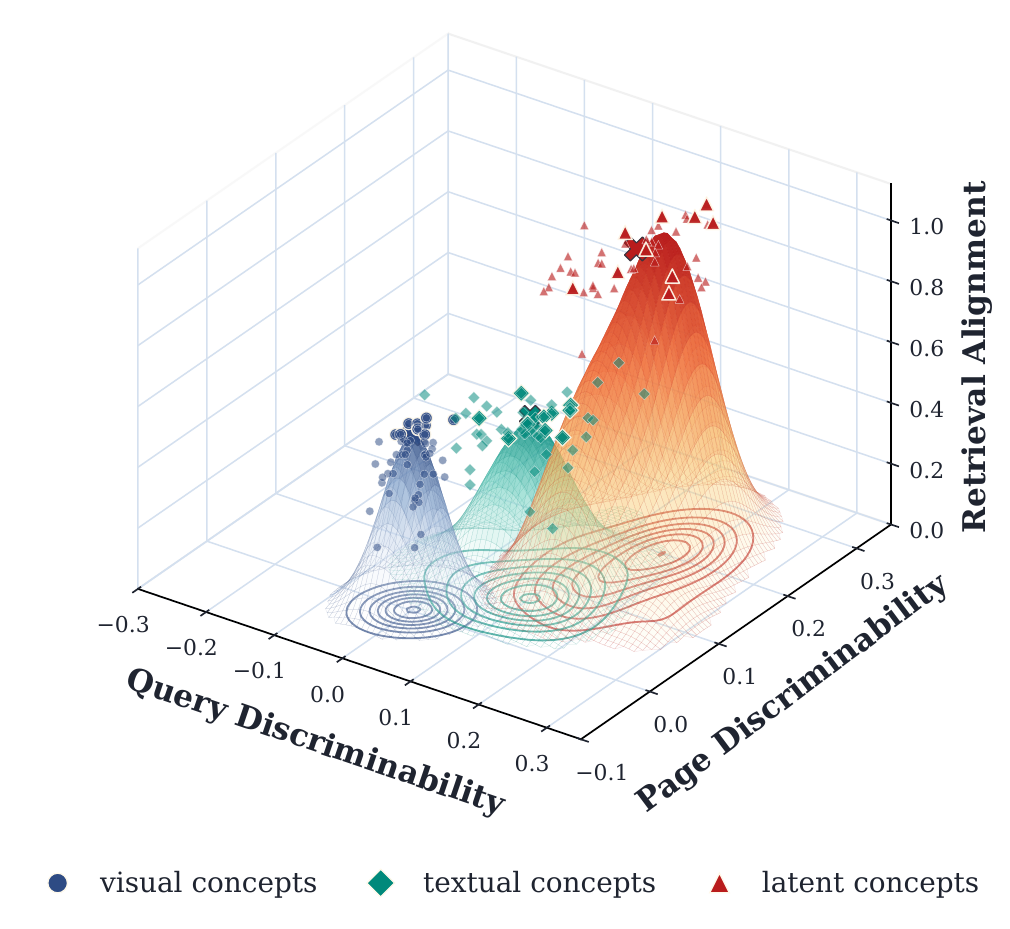}
        \caption{ChartQA.}
        % \label{fig:font_text_form}
    \end{subfigure}\hfill%
    \begin{subfigure}[t]{0.48\linewidth}
        \centering
        \includegraphics[width=\linewidth]{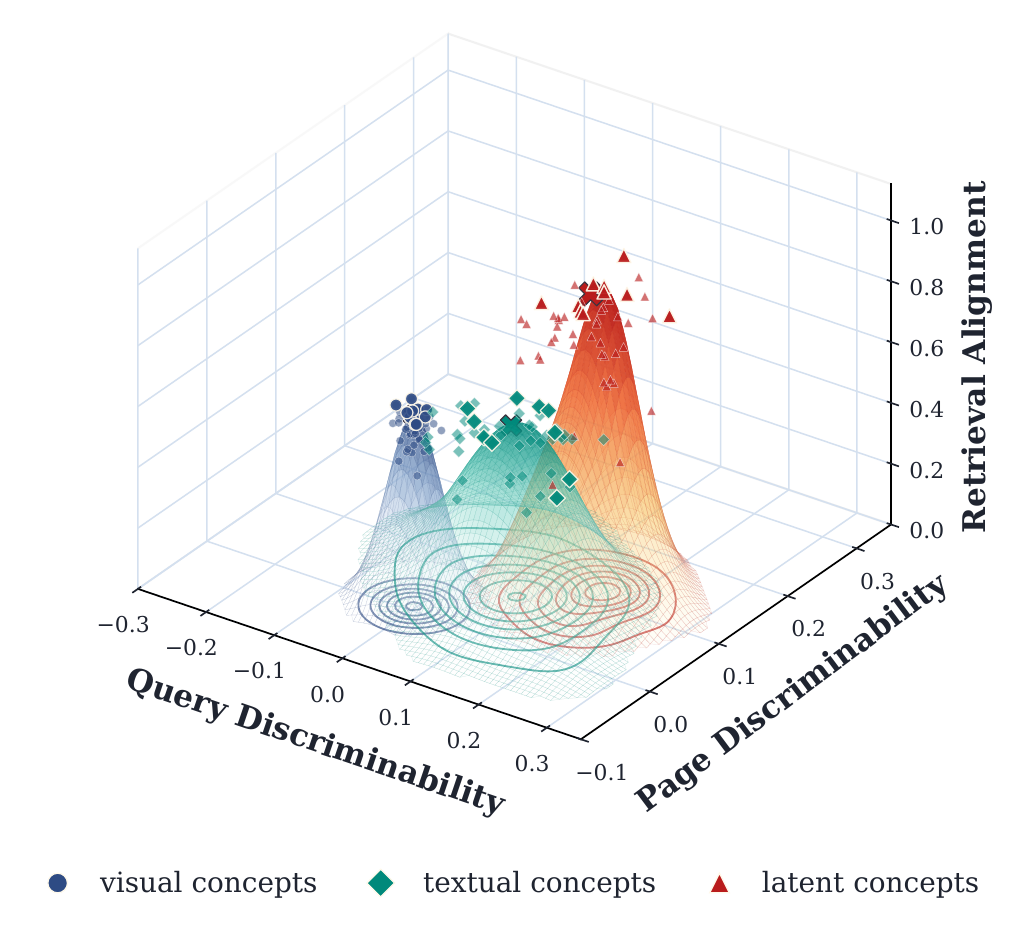}
        \caption{SlideVQA.}
        % \label{fig:layout_structure}
    \end{subfigure}\hfill%
    \begin{subfigure}[t]{0.48\linewidth}
        \centering
        \includegraphics[width=\linewidth]{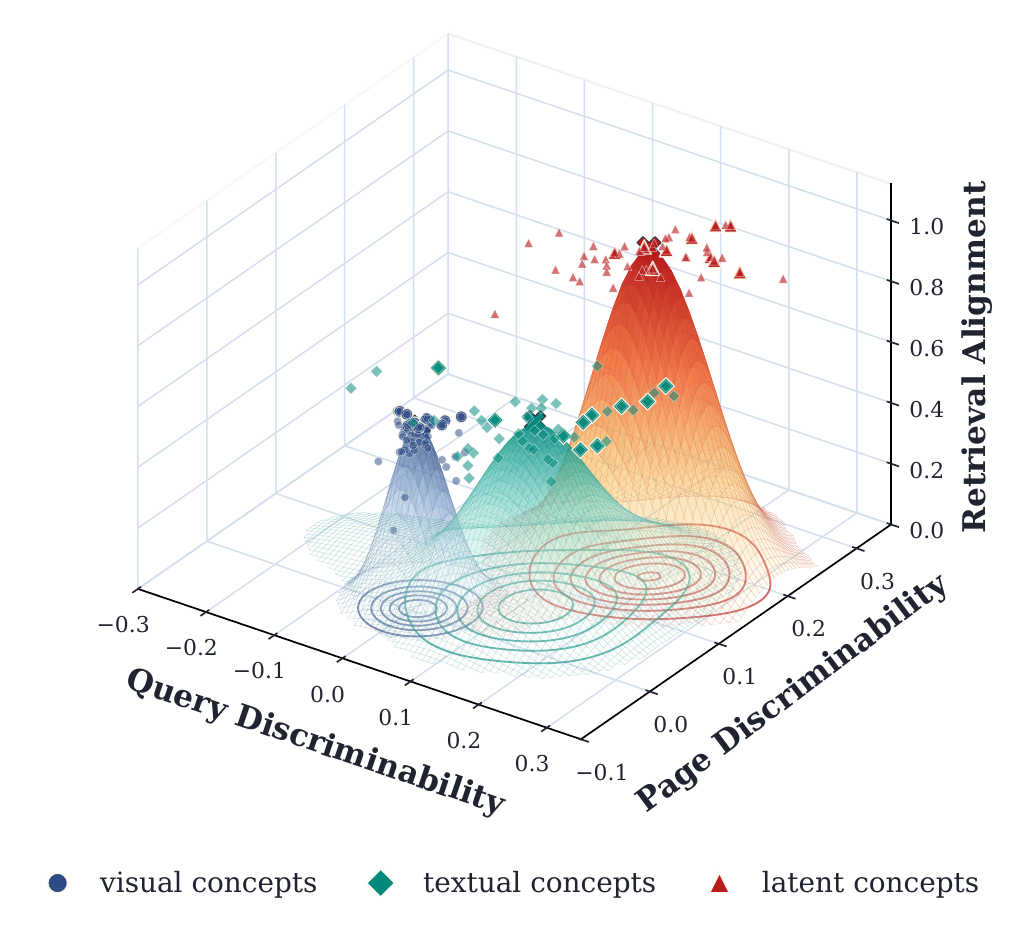}
        \caption{InfoVQA.}
        % \label{fig:mixed_typesetting}
    \end{subfigure}\hfill%
    \begin{subfigure}[t]{0.48\linewidth}
        \centering
        \includegraphics[width=\linewidth]{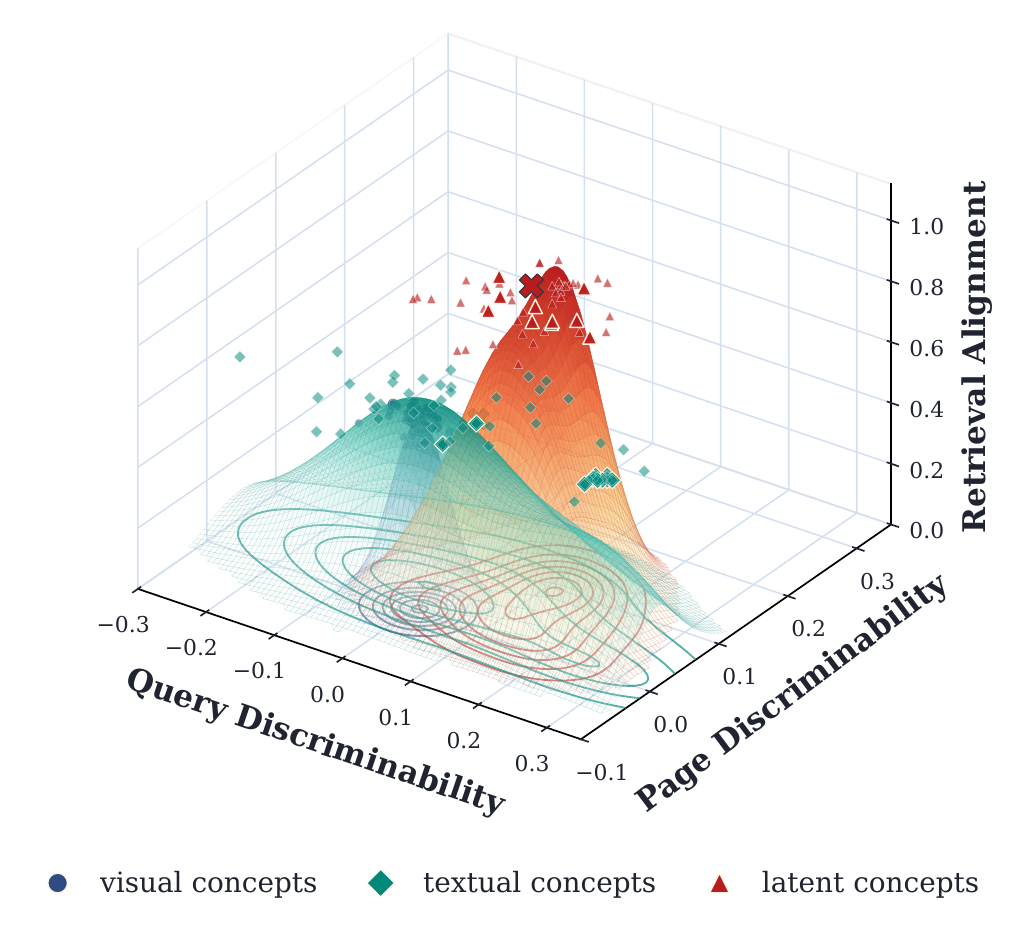}
        \caption{Wikimedia Maps.}
        % \label{fig:natural_interference}
    \end{subfigure}\hfill%
    \caption{3D retrieval geometry of concept representations. We compare visual and textual concept proxies with \method{} latent concepts using query discriminability, page discriminability, and retrieval alignment. }
    \label{fig:alignment}
    \vspace{-0.2in}
\end{figure}
\begin{figure}[t]
    \centering
    \begin{subfigure}[t]{0.48\linewidth}
        \centering
        \includegraphics[width=\linewidth]{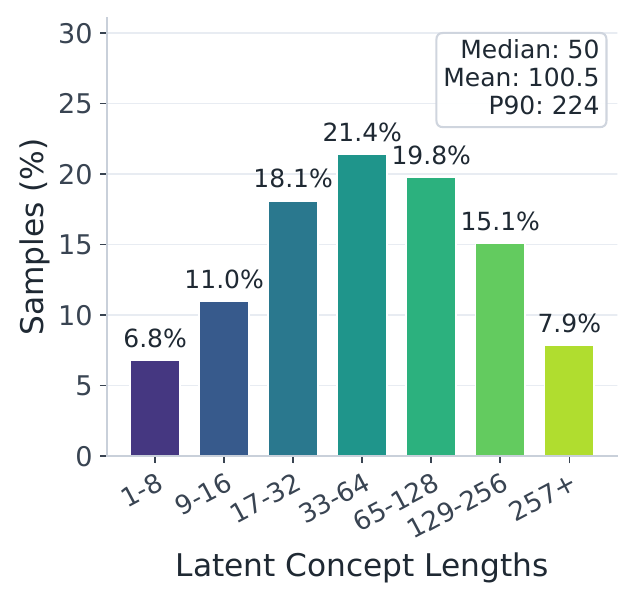}
        \caption{Length distribution of allocated latent tokens.}
        \label{fig:distribution}
    \end{subfigure}\hfill%
    \begin{subfigure}[t]{0.48\linewidth}
        \centering
        \includegraphics[width=\linewidth]{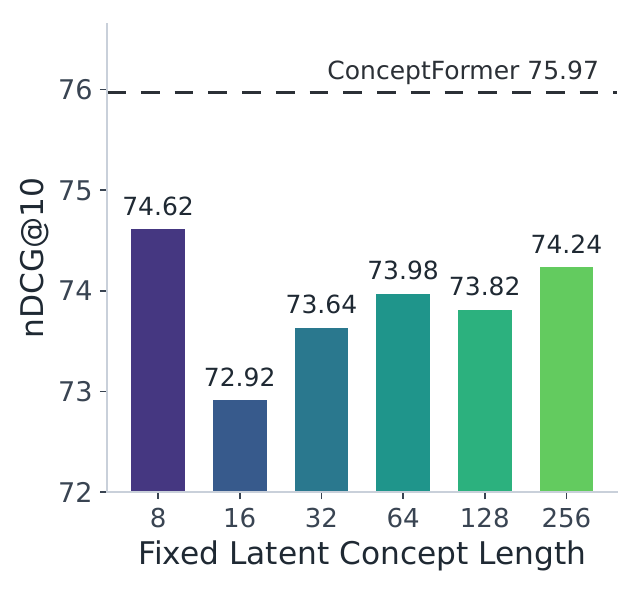}
        \caption{Performance under fixed latent token lengths.}
        \label{fig:performance}
    \end{subfigure}\hfill%
    \vspace{-0.1in}
    \caption{Effectiveness of adaptive latent concept of \method{}. All models are implemented using Qwen2.5-VL, where P90 denotes the 90th-percentile length. }
    \label{fig:capacity_analysis}
    % \vspace{-0.2in}
\end{figure}
\begin{figure*}[t]
    \centering
    \includegraphics[width=1.0\linewidth]{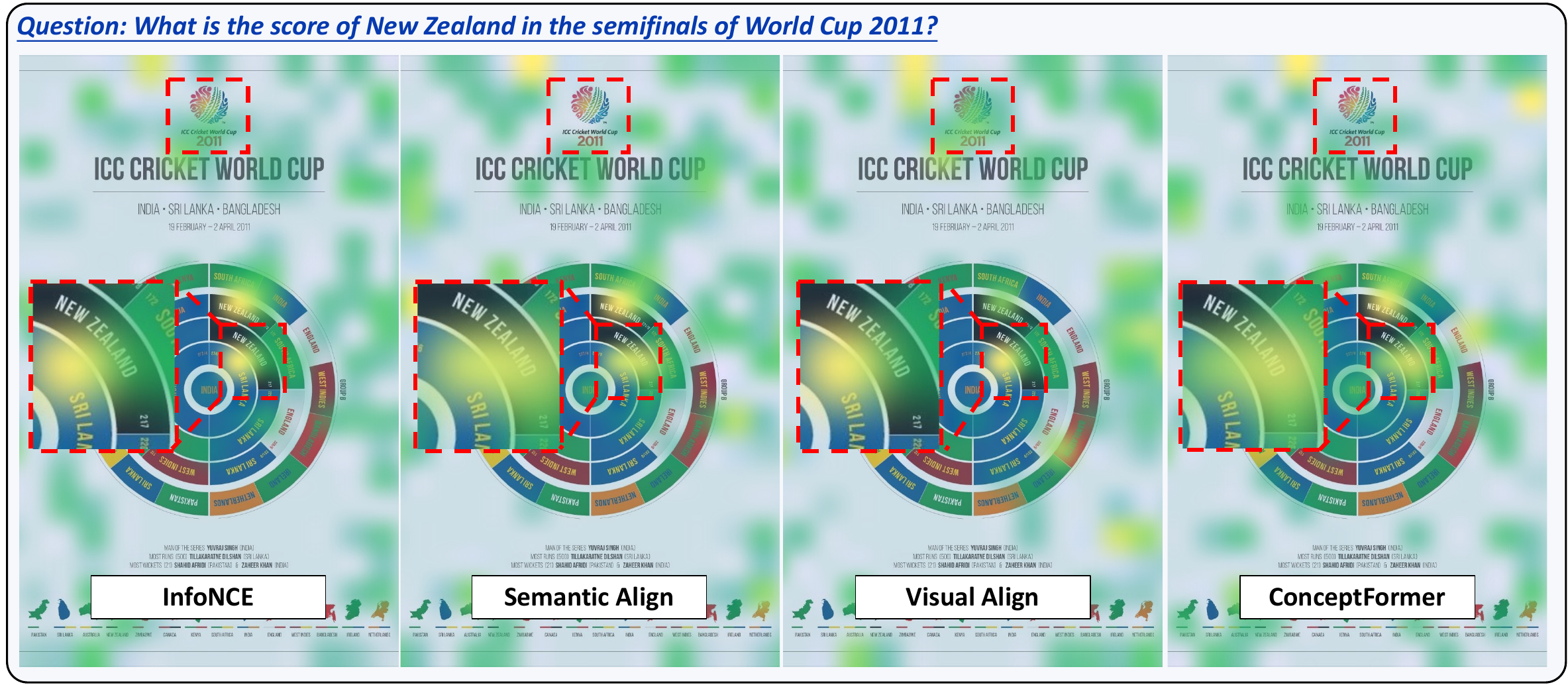}
    \caption{Case study. Regions with higher color intensity indicate stronger patch-level query relevance.}
    \label{fig:case_study}
    \vspace{-0.1in}
\end{figure*}
\subsection{Ablation Study}
We conduct ablation studies to examine the key designs of latent concept representation learning in \method{}. Table~\ref{tab:ablation_align} compares alternative concept representation methods,  concept representation strategies, and auxiliary feature-level grounding losses.

\textbf{Effectiveness of Different Latent Concept Representations.}
We compare \method{} with two alternative concept representations derived from query-related evidence: a textual concept proxy (\method{} (Text)) and a visual concept proxy (\method{} (Vision)). The textual proxy represents latent concepts using the textual cues extracted from the evidence proposals and is trained by aligning the resulting document relevance distribution with the query-induced document distribution. The visual proxy instead represents latent concepts by aggregating the visual tokens corresponding to the evidence regions, obtained by projecting the evidence boxes onto the retriever's visual token grid, and is optimized with the same supervision objective.

As shown in the evaluation results, \method{} consistently outperforms both \method{} (Text) and \method{} (Vision), demonstrating the effectiveness of learning query-related evidence through latent concept representations. Furthermore, \method{} (Text) achieves substantially better performance than \method{} (Vision), suggesting that textual verbalization provides more informative semantic supervision than directly aggregating visual features for aligning queries with visual documents.

\textbf{Relevance Signals Distributed across Latent Concepts.}
We further investigate how latent concepts should be supervised by comparing our Query-Concept Matching objective with a contrastive alternative. Specifically, the Query-Concept Matching objective aligns the query-induced relevance distribution over candidate documents with the relevance distribution induced by matching the query against the latent concepts extracted from those documents. We also evaluate a variant, \method{} w/o Mean Pooling, which removes mean pooling over all latent concept tokens and instead uses only the hidden state of the end token of latent concept ``\texttt{<|lcon\_end|>}'' as the concept representation.

Compared with the contrastive Query-Concept Matching baseline, \method{} achieves consistently better performance, indicating that aligning relevance distributions provides a more effective supervision signal for learning latent concepts than conventional contrastive training. Moreover, \method{} outperforms \method{} w/o Mean Pooling, demonstrating that query-relevant information is distributed across multiple latent concept states rather than being fully captured by a single hidden state. This finding supports our design of adaptive latent concepts as a distributed concept representation, where different latent concept states encode complementary local evidence and cross-region interactions required for accurate query-document matching.

\textbf{Sensitivity to the Representation-Loss Weight.}
% \label{app:lambda}
We study the effect of the latent concept representation-loss weight $\lambda$ in Eq.~\ref{eq:loss}. When $\lambda=0$, the concept-level ranking supervision is removed, reducing the objective to conventional page-level contrastive learning. As shown in Table~\ref{tab:ablation_lambda}, the best average performance on the Qwen2.5-VL backbone is achieved with $\lambda=0.2$. Increasing $\lambda$ beyond this point does not consistently improve retrieval performance, suggesting that overly strong distribution alignment may compromise the discriminative signals required for query--page matching.
% \textbf{Training Strategies with Latent Concept Supervision.}
% We further investigate different strategies for incorporating latent concept supervision during training. We first construct \method{} w/o KL Loss by removing the alignment loss $\mathcal{L}_{\mathrm{align}}$, leaving the model trained solely with the standard InfoNCE objective~\cite{oord2018representation} for page-level contrastive learning. Based on this variant, we further replace $\mathcal{L}_{\mathrm{align}}$ with feature-level supervision using either Mean Squared Error (MSE) loss or cosine similarity loss to align the latent concept representations during training.

% Compared with \method{}, \method{} w/o KL Loss suffers a noticeable performance degradation, indicating that relying solely on query-document relevance supervision is insufficient for learning meaningful relevance representations. Without latent concept guidance, the retriever is not explicitly encouraged to capture the fine-grained evidence that explains why a query should rank one candidate page ahead of another. Furthermore, although the variants trained with MSE or cosine similarity losses yield improvements, they fail to surpass \method{} in overall performance. These results suggest that latent concepts should not be constrained to reconstruct local visual features. Instead, \method{} uses a KL-divergence objective to directly align latent concept learning with the retrieval objective, enabling latent concepts to bridge the gap between localized visual evidence and page-level query--document relevance.

\subsection{Embedding Space Visualization of Learned Latent Concepts}

To characterize the retrieval properties of different concept representations, we compare the visual concept proxy, the textual concept proxy, and the latent concept representation learned by \method{} from three complementary perspectives. Query Discriminability measures how well a representation separates its matched query from mismatched queries, while Page Discriminability evaluates its ability to distinguish the positive page from negative pages. Retrieval Alignment further measures how closely the document-ranking distribution induced by a representation matches that of the original query. The first two metrics quantify pairwise discrimination, whereas Retrieval Alignment captures consistency over the entire retrieval ranking. Higher values indicate stronger query discrimination, page discrimination, and retrieval alignment, respectively. All representations are evaluated using the same negative queries, negative pages, and candidate sets. More experimental details are provided in Appendix~\ref{app:retrieval_geometry}.

Figure~\ref{fig:alignment} shows that visual concept proxies preserve fine-grained visual grounding but provide relatively weak query/page discrimination and retrieval alignment. Textual concept proxies capture higher-level semantic abstractions, yet their performance degrades on image-intensive benchmarks such as Wikimedia Maps, where geographic layouts, spatial relationships, color-coded regions, and distributed textual labels cannot be faithfully represented through textual descriptions alone. In contrast, the latent concepts learned by \method{} consistently achieve stronger Query Discriminability, Page Discriminability, and Retrieval Alignment. These results suggest that the learned latent concepts effectively integrate localized visual evidence, query semantics, and page-level contextual signals into a unified retrieval-oriented representation, rather than merely imitating either visual or textual concept proxies.

\subsection{Analyses of Token Allocation Behaviors for Latent Concepts of \method{}}
To better understand the effectiveness of adaptive concept capacity, we analyze the token allocation behavior of \method{} and compare it with variants using fix token numbers of latent concepts, as shown in Figure~\ref{fig:capacity_analysis}.

In Figure~\ref{fig:distribution}, we first present the distribution of latent concept lengths, estimated from the visual token coverage of query-related evidence regions. The allocated lengths vary substantially across samples. More than half of the samples require no more than 64 latent concept tokens, with a median length of 50. Meanwhile, the distribution exhibits a pronounced long tail: 15.1\% of the samples require 129-256 tokens, 7.9\% require more than 256 tokens, and the P90 length reaches 224. These observations suggest that visual document retrieval exhibits highly diverse concept-capacity requirements rather than a single fixed capacity.
Then we compare retrieval performance under different fixed latent concept tokens in Figure~\ref{fig:performance}. All variants use the same training objective and representation learning strategy as \method{}, differing only in whether the latent concept length is fixed or adaptively allocated. Increasing the latent concept length does not yield monotonic improvements, indicating that the performance gains cannot be attributed simply to using more latent tokens. Instead, \method{} consistently outperforms all fixed-length variants in terms of NDCG@10. This result suggests that adapting concept capacity to each query--document pair is more effective than assigning a uniform latent length, as it avoids allocating redundant concept states to simple examples while preserving sufficient representational capacity for visually complex or spatially dispersed evidence.

\subsection{Case Study}
Figure~\ref{fig:case_study} provides a qualitative comparison of patch-level retrieval signals. 
We visualize the normalized patch scores over the page, where higher color intensity indicates stronger contribution to query-page matching in the retriever's representation space.

For the query ``What is the score of New Zealand in the semifinals of World Cup 2011?'', the model needs to combine two types of evidence. The top region provides the global event context, namely ``ICC Cricket World Cup 2011'', while the central circular chart contains the fine-grained local evidence. The InfoNCE-only model activates several broadly related or visually salient regions, such as the enlarged New Zealand area, parts of the circular chart, and peripheral page elements, but it does not clearly connect the event-year context with the small score evidence required by the question. Textual alignment better responds to the local textual clue around New Zealand, showing the benefit of language-expressible evidence descriptions. However, it gives a weaker response to the top ``World Cup 2011'' event context, suggesting that textual evidence alone may emphasize local lexical cues while missing broader page-level visual context. Visual alignment shows the opposite pattern: it captures the logo and year more clearly but remains weak on the fine-grained chart score, suggesting that visual grounding alone is insufficient for small query-critical numbers.
In contrast, \method{} produces a more balanced retrieval signal, responding to both the top event-year region and the central chart region containing New Zealand’s semifinal score. This indicates that the latent concepts help the retriever bind page-level context with local numerical evidence, rather than treating them as isolated visual or textual cues. 
% By converting both regions into a coherent patch-level matching signal, \method{} forms a more evidence-aware page representation for visual document retrieval.
\section{Conclusion}
% In this paper, we propose \method{}, a latent concept representation learning framework for visual document retrieval. \method{} introduces adaptive latent concepts during training to represent query-conditioned relevance signals, bridging localized visual grounding and semantic relevance through representation learning. By dynamically allocating concept capacity according to query-related evidence coverage, \method{} captures fine-grained information while preserving the standard retrieval pipeline during inference.
% Experiments on diverse visual document retrieval benchmarks demonstrate that \method{} consistently outperforms baseline retrievers. Further analyses show that the learned latent concepts achieve stronger retrieval alignment than textual and visual concept proxies, and adaptive concept capacity better matches the varying evidence scale of visually rich documents.
In this paper, we propose \textbf{\method{}}, a latent concept learning framework for visual document retrieval. \method{} introduces adaptive latent concepts to encode query-conditioned relevance, bridging localized visual grounding and semantic relevance. By dynamically allocating concept capacity according to evidence coverage, it captures fine-grained information while retaining the standard retrieval pipeline at inference. Experiments across diverse benchmarks show consistent improvements over baseline retrievers. Further analyses demonstrate stronger retrieval alignment than textual and visual concept proxies, while adaptive capacity better accommodates varying evidence scales in visually rich documents.

% \begin{acks}

% \end{acks}
\bibliographystyle{ACM-Reference-Format}
% \balance
\bibliography{sample-base}
\appendix
\clearpage
\newpage
\section{Additional Experimental Details}
\label{app:details}

\subsection{Datasets and Evaluation Tasks}
\label{app:datasets}

\paragraph{Training Data.}
We follow the data construction protocol described in Section~4 and train \method{} on approximately 38K query-document pairs collected from DocVQA, InfoVQA, VisualMRC, OpenWikiTable, DUDE, and MHDocVQA. For each positive query-page pair, we additionally generate query-related region proposals for adaptive concept capacity allocation. The processed training set is stored as a merged JSONL file containing the query, positive page, bounding-box annotations, and corresponding textual cues.  Dataset statistics are reported in Table~\ref{tab:training_dataset_statistics}

\paragraph{Evaluation Benchmarks.}
\begin{table}[t]
\small
\centering
\caption{Training dataset statistics.}
\label{tab:training_dataset_statistics}
\begin{tabular}{llrr}
\toprule
Dataset & Field & \#Images & \#Queries \\
\midrule
DocVQA        & Industry    & 12,767 & 6,382 \\
InfoVQA       & Infographic & 5,485  & 9,592 \\
VisualMRC     & Webpage     & 10,229 & 6,126 \\
OpenWikiTable & Table       & 1,257  & 4,261 \\
DUDE          & Open        & 27,955 & 2,135 \\
MHDocVQA      & Open        & 28,550 & 9,470 \\
\bottomrule
\end{tabular}
\end{table}
\begin{table}[t]
\small
\centering
\caption{Test dataset statistics.}
\label{tab:test_dataset_statistics}
\begin{tabular}{llrr}
\toprule
Dataset & Field & \#Images & \#Queries \\
\midrule
InfoVQA          & Infographic & 5,485  & 1,048 \\
ChartQA          & Chart       & 20,882 & 150 \\
SlideVQA         & Slide       & 52,380 & 760 \\
TQA              & Table       & 1,000  & 1,000 \\
OWID Charts      & Chart       & 131    & 131 \\
Wikimedia Maps   & Map         & 455    & 455 \\
\bottomrule
\end{tabular}
\end{table}
We evaluate \method{} on six visual document retrieval benchmarks covering different document formats and evidence structures. InfoVQA requires the retriever to identify an infographic that answers a natural-language question. ChartQA and SlideVQA respectively retrieve relevant chart and presentation-slide images. TQA retrieves textbook pages containing the evidence required by the question. OWID Charts evaluates retrieval over statistical charts from Our World in Data, while Wikimedia Maps focuses on visually grounded geographic information in map images.  Benchmark statistics are reported in Table~\ref{tab:test_dataset_statistics}

\begin{table}[t]
\centering
\small
\caption{Sources of the six evaluation benchmarks. All identifiers refer to
Hugging Face datasets.}
\label{tab:dataset_sources}
\begin{tabular}{p{0.25\columnwidth}p{0.67\columnwidth}}
\toprule
\textbf{Benchmark} & \textbf{Dataset Identifier} \\
\midrule
InfoVQA &
\texttt{NTT-hil-insight/OpenDocVQA};
images from \texttt{OpenDocVQA-Corpus} \\
ChartQA &
\texttt{NTT-hil-insight/OpenDocVQA};
images from \texttt{OpenDocVQA-Corpus} \\
SlideVQA &
\texttt{NTT-hil-insight/OpenDocVQA};
images from \texttt{OpenDocVQA-Corpus} \\
TQA &
\texttt{jinaai/tqa} \\
OWID Charts &
\texttt{jinaai/owid\_charts\_en} \\
Wikimedia Maps &
\texttt{jinaai/wikimedia-commons-maps} \\
\bottomrule
\end{tabular}
\end{table}
\begin{table}[ht]
\centering
\small
\caption{Main training and inference configurations of \method{}.}
\label{tab:training_config}
\begin{tabular}{p{0.45\columnwidth}p{0.45\columnwidth}}
\toprule
\textbf{Configuration} & \textbf{Value} \\
\midrule
Main backbone & Qwen2.5-VL-7B-Instruct \\
Alternative backbone & Phi3V / DSE initialization \\
Training epochs & 3 \\
Optimizer & AdamW \\
Learning rate & $1\times10^{-4}$ \\
Per-device batch size & 8 \\
Gradient accumulation & 4 \\
Number of GPUs & 4 \\
Effective batch size & 128 \\
Precision & BF16 \\
LoRA rank / alpha / dropout & 8 / 64 / 0.1 \\
Retrieval readout & EOS representation \\
Representation normalization & $\ell_2$ normalization \\
Retrieval temperature & $\tau=0.01$ \\
Concept-state aggregation & Mean pooling \\
Representation-loss weight & $\lambda=0.2$ \\
Additional textual KL weight & 0 \\
Concept length & Dynamically allocated \\
\bottomrule
\end{tabular}
\end{table}
\begin{table*}[t]
\centering
\small
\caption{Public checkpoints used for baseline evaluation. All identifiers
refer to Hugging Face model repositories.}
\label{tab:baseline_checkpoints}
\begin{tabular}{p{0.18\linewidth}p{0.22\linewidth}p{0.52\linewidth}}
\toprule
\textbf{Category} & \textbf{Method} & \textbf{Checkpoint Identifier} \\
\midrule
\multirow{4}{*}{Text retrievers}
& BM25 & No model checkpoint \\
& BGE-1.5 & \texttt{BAAI/bge-large-en-v1.5} \\
& E5-Mistral & \texttt{intfloat/e5-mistral-7b-instruct} \\
& NV-Embed-v2 & \texttt{nvidia/NV-Embed-v2} \\
\midrule
\multirow{6}{*}{Multimodal retrievers}
& CLIP & \texttt{openai/clip-vit-large-patch14-336} \\
& E5-V & \texttt{royokong/e5-v} \\
& Jina-CLIP-v2 & \texttt{jinaai/jina-clip-v2} \\
& VLM2Vec & \texttt{TIGER-Lab/VLM2Vec-Qwen2VL-2B} \\
\midrule
\multirow{4}{*}{Visual document retrievers}
& ColPali &
\texttt{vidore/colpali}, initialized from
\texttt{vidore/colpaligemma-3b-mix-448-base} \\
& DSE & \texttt{Tevatron/dse-phi3-docmatix-v1} \\
& VDocRetriever &
\texttt{NTT-hil-insight/VDocRetriever-Phi3-vision} \\
& VisRAG-Ret & \texttt{openbmb/VisRAG-Ret} \\
\bottomrule
\end{tabular}
\end{table*}
\subsection{Baseline Checkpoints}
\label{app:baselines}
For OCR-based retrieval, we first extract page text with PaddleOCR and then apply the corresponding sparse or dense text retriever. Visual and multimodal retrievers directly encode the original page images. We use publicly released checkpoints and follow the official preprocessing and inference settings of each method whenever available. Table~\ref{tab:baseline_checkpoints} lists the checkpoints used in the main and supplementary experiments.

\subsection{Implementation Details}
\label{app:implementation}
Table~\ref{tab:training_config} shows all the implementation details.
\paragraph{Retriever Training.}
The main retriever is initialized from Qwen2.5-VL-7B-Instruct. For the Phi3V setting, we initialize the retriever from Tevatron/dse-phi3-docmatix-v1 and use VDocRetriever-Phi3-vision as the LoRA initialization. All \method{} variants are trained for three epochs with AdamW. We use a per-device batch size of 8, gradient accumulation over 4 steps, and 4 GPUs, resulting in an effective batch size of 128.

We perform parameter-efficient fine-tuning with LoRA. The LoRA rank is set to 8, the scaling factor is 64, and the dropout rate is 0.1. LoRA is applied to \texttt{q\_proj}, \texttt{k\_proj}, \texttt{v\_proj}, \texttt{o\_proj}, \texttt{down\_proj}, \texttt{up\_proj}, and \texttt{gate\_proj}. Training uses BF16 precision and a learning rate of $1\times10^{-4}$.

\paragraph{Retrieval and Concept Representations.}
Query and page representations are read from the EOS position, followed by $\ell_2$ normalization. Retrieval probabilities are computed with a temperature of $0.01$. The latent concept representation is obtained by mean pooling over all latent concept states. The representation objective uses the forward KL direction,
\begin{equation}
\mathrm{KL}\!\left(
P(d\mid q_i,\widetilde{\mathcal D}_i)
\,\|\,  P(d\mid\hat c_i,\widetilde{\mathcal D}_i) \right), 
\end{equation}
with $\lambda=0.2$. We do not apply an additional textual-distribution KL objective. No fixed concept-token budget is imposed in the full model; the latent concept length is determined dynamically from the query-related bounding boxes.

\subsection{Evidence Proposal and Concept Capacity Allocation}
\label{app:evidence_proposal}

\paragraph{Evidence Proposer.}
\begin{figure}[t] 
\centering
\begin{tcolorbox}[promptbox, title=System Prompt for Evidence Extraction.]
\textbf{System Prompt:} \\[2pt]
You are a visual grounding assistant.

Task: Given an image and a question, think step by step to find regions containing all evidence needed to answer. Each region must be self-contained--able to answer the query on its own. When unsure, use larger boxes to ensure completeness and readability.

Region-selection guidelines:
1. Fully cover key evidence plus immediate context; do not clip text, numbers, or symbols.
2. Prefer complete information units (full words/lines; entire signs/labels; for charts include legend, axes, units, titles/notes).
3. Tables: include the header and relevant rows/columns with necessary context; avoid single-cell crops.
4. If evidence spans multiple parts, use multiple boxes--or one larger box if they're adjacent.
5. Images/illustrations: include nearby numeric values or captions required by the question.

Output format:
[ {"bbox\_2d": [x1, y1, x2, y2], "label": "a short description of this region and why it is relevant"}] \\[6pt]
\tcblower
\textbf{User Prompt:} \\[2pt]
Query: {\{Query\}} \\[2pt]
Images: {\{Positive Image\}}
\end{tcolorbox}
\caption{Prompt templates used in Evidence Extraction.}
\label{fig:prompt}
\end{figure}
We use Qwen3.6-Plus as the training-time evidence proposer. The proposer
receives the query and the positive page image, and returns one or more
query-related regions. We use a decoding temperature of $0.2$ and process
requests with 16 concurrent workers. The proposer is used only for constructing
training supervision and is not invoked during retrieval inference.

Figure~\ref{fig:prompt} shows the system and user prompt, which is designed to encourage complete and readable evidence
regions.
The bounding box supplies the spatial grounding used for concept-capacity
allocation. The corresponding textual label is retained as a semantic
description and is used to construct the textual concept proxy in the
ablation study.

\paragraph{Mapping Regions to the Visual Token Grid.}
For each proposal
$b_{i,n}=(x^1_{i,n},y^1_{i,n},x^2_{i,n},y^2_{i,n})$, we map its coordinates
onto the visual token grid produced by the retriever and collect the covered
token indices:
\begin{equation}
\mathcal I_{i,n}
=
\operatorname{Index}_{\pi_\theta}
(d_i^+,b_{i,n}).
\end{equation}
The latent concept length is computed as
\begin{equation}
T_i
=
\sum_{n=1}^{N_i}
|\mathcal I_{i,n}|.
\end{equation}
Thus, the concept capacity is tied to the retriever-specific coverage of the
query-related regions. Samples with localized evidence receive shorter latent
concept sequences, whereas samples involving larger charts, maps, tables, or
cross-region relations receive greater concept capacity. The setting
``fixed latent tokens = 0'' in our implementation denotes this dynamic
bbox-based allocation rather than an empty concept sequence.

\subsection{Sensitivity to the Representation-Loss Weight}
\label{app:lambda}
We investigate the effect of the latent concept representation-loss weight
$\lambda$ in Eq.~\ref{eq:loss}.
Setting $\lambda=0$ removes concept-side ranking supervision and reduces the
training objective to standard page-level contrastive learning. A positive
$\lambda$ encourages the latent concept representation to reproduce the
query-induced ranking behavior over candidate pages.

Table~\ref{tab:ablation_lambda} shows that $\lambda=0.2$ gives the best average result
on the main Qwen2.5-VL backbone. Performance is not monotonic as $\lambda$
increases, indicating that overly strong distribution matching can interfere
with the discriminative query-page objective. Results on Phi3V are less
sensitive to $\lambda$: values from $0.1$ to $0.5$ remain close, and the
numerically best result is obtained at $\lambda=0.5$. We use $\lambda=0.2$
as a common default because it is optimal for the main backbone and remains
near the optimum for Phi3V.

\section{Retrieval Geometry Metrics}
\label{app:retrieval_geometry}

For each sample $i$, we compare the visual concept proxy $\hat{\mathbf{v}}_i$, the textual concept proxy $\hat{\mathbf{r}}_i$, and the latent concept representation $\hat{\mathbf{c}}_i$ produced by \method{}. Given a representation
$\mathbf{a}_i \in
\{\hat{\mathbf{v}}_i,\hat{\mathbf{r}}_i,\hat{\mathbf{c}}_i\}$,
We define Query Discriminability as
\begin{equation}
D_q(\mathbf{a}_i)
=
\operatorname{sim}(\mathbf{a}_i,\mathbf{q}_i)
-
\frac{1}{|\mathcal{Q}_i^-|}
\sum_{\mathbf{q}_j \in \mathcal{Q}_i^-}
\operatorname{sim}(\mathbf{a}_i,\mathbf{q}_j),
\label{eq:query_discriminability}
\end{equation}
where $\mathcal{Q}_i^-$ contains mismatched queries. A larger
$D_q(\mathbf{a}_i)$ indicates that the representation more clearly
distinguishes the matched query from unrelated queries.

Page Discriminability is defined as
\begin{equation}
D_p(\mathbf{a}_i)
=
\operatorname{sim}(\mathbf{a}_i,\mathbf{d}_i^+)
-
\frac{1}{|\mathcal{D}_i^-|}
\sum_{\mathbf{d}_j \in \mathcal{D}_i^-}
\operatorname{sim}(\mathbf{a}_i,\mathbf{d}_j),
\label{eq:page_discriminability}
\end{equation}
where $\mathbf{d}_i^+$ is the positive page representation and
$\mathcal{D}_i^-$ contains negative pages. A larger
$D_p(\mathbf{a}_i)$ indicates stronger separation between the
positive page and irrelevant pages.

We further define Retrieval Alignment as
\begin{equation}
R(\mathbf{a}_i)
=
\exp\left(
-\operatorname{KL}
\left(
P(d\mid q_i,\widetilde{\mathcal{D}}_i)
\parallel
P(d\mid \mathbf{a}_i,\widetilde{\mathcal{D}}_i)
\right)
\right),
\label{eq:retrieval_alignment}
\end{equation}
where a higher $R(\mathbf{a}_i)$ indicates that the representation
induces a document-ranking distribution closer to that of the
original query. We use the same $\mathcal{Q}_i^-$,
$\mathcal{D}_i^-$, and $\widetilde{\mathcal{D}}_i$ for all three
representation types to ensure a consistent comparison.
% \input{section-sigir/appendix}

\end{document}